\documentclass[letterpaper, 10 pt, journal, twoside]{IEEEtran}  % Comment this line out if you need a4paper

\IEEEoverridecommandlockouts                              % This command is only needed if 
\makeatletter
\let\NAT@parse\undefined
\makeatother

\newcommand\copyrighttext{%
  \footnotesize \textcopyright 2021 IEEE. Personal use of this material is permitted.
  Permission from IEEE must be obtained for all other uses, in any current or future
  media, including reprinting/republishing this material for advertising or promotional
  purposes, creating new collective works, for resale or redistribution to servers or
  lists, or reuse of any copyrighted component of this work in other works.}
\newcommand\copyrightnotice{%
\begin{tikzpicture}[remember picture,overlay]
\node[anchor=south,yshift=2pt] at (current page.south) {\fbox{\parbox{\dimexpr\textwidth-\fboxsep-\fboxrule\relax}{\copyrighttext}}};
\end{tikzpicture}%
}

\usepackage[usenames, dvipsnames, table]{xcolor}
\let\labelindent\relax
\usepackage{enumitem}
\usepackage[ruled,linesnumbered,resetcount,vlined]{algorithm2e}
\SetKwComment{Comment}{/* }{ */}
\SetKwComment{SmallComment}{//}{}
\SetKwFor{For}{for (}{)}{}
\usepackage{graphicx} % for pdf, bitmapped graphics files
\usepackage{caption}
\usepackage{subcaption}
\usepackage{multicol}
\usepackage{amsmath}
\usepackage{amssymb}
\usepackage{float}
\usepackage{multirow}
\usepackage{tabularx}

\usepackage[markup=bfit, deletedmarkup=sout, authormarkup=superscript, commandnameprefix=ifneeded]{changes}
\definechangesauthor[name={td}, color=      red] {TODO}
\definechangesauthor[name={bh}, color= NavyBlue] {BH}
\definechangesauthor[name={ar}, color=PineGreen] {AR}
\definechangesauthor[name={ap}, color=Purple] {AP}
\definechangesauthor[name={yt}, color=Orange] {YT}
\definechangesauthor[name={jk}, color=RubineRed] {JK}

\usepackage{subcaption}
\usepackage[]{hyperref}
\colorlet{mylinkcolor}{black}
\colorlet{mycitecolor}{black}
\colorlet{myurlcolor}{black}

\hypersetup{
  linkcolor  = mylinkcolor,
  citecolor  = mycitecolor,
  urlcolor   = myurlcolor,
  colorlinks = true,
}

\usepackage[capitalize,noabbrev]{cleveref}

\begin{document}

\title{PokeRRT: Poking as a Skill and Failure Recovery Tactic for Planar Non-Prehensile Manipulation}

\author{Anuj Pasricha$^{*}$, Yi-Shiuan Tung, Bradley Hayes, and Alessandro Roncone%
\thanks{Manuscript received: September 9, 2021; Revised: December 11, 2021; Accepted: January 14, 2022.}
\thanks{This paper was recommended for publication by Editor Stephen J. Guy upon evaluation of the Associate Editor and Reviewers' comments.}
\thanks{The authors are with the Department of Computer Science,
	     University of Colorado Boulder, 1111 Engineering Drive, Boulder, CO USA
        {\tt\footnotesize firstname.lastname@colorado.edu}}%
\thanks{$*$ Corresponding author.}%
\thanks{Digital Object Identifier (DOI): see top of this page.}
}

\markboth{IEEE Robotics and Automation Letters. Preprint Version. Accepted January, 2022}
{Pasricha \MakeLowercase{\textit{et al.}}: P\MakeLowercase{oke}RRT: Poking as a Skill and Failure Recovery Tactic for Planar Non-Prehensile Manipulation}

\maketitle
\copyrightnotice

\begin{abstract}
% \jk{Y'all are all from the same place, so no need for the '1' superscript in the author list. }
In this work, we introduce \textsl{PokeRRT}, a novel motion planning algorithm that demonstrates poking as an effective non-prehensile manipulation skill to enable fast manipulation of objects and increase the size of a robot's reachable workspace. We showcase poking as a failure recovery tactic used synergistically with pick-and-place for resiliency in cases where pick-and-place initially fails or is unachievable. Our experiments demonstrate the efficiency of the proposed framework in planning object trajectories using poking manipulation in uncluttered and cluttered environments. In addition to quantitatively and qualitatively demonstrating the adaptability of \textsl{PokeRRT} to different scenarios in both simulation and real-world settings, our results show the advantages of poking over pushing and grasping in terms of success rate and task time.%
\end{abstract}

\section{Introduction}\label{sec:introduction}

\IEEEPARstart{H}{umans} engage naturally in multiple forms of dexterous manipulation that involve grasping, pushing, poking, rolling, and tossing objects \cite{bullock2011classifying}.
Consequently, the development of similar functionality in autonomous machines is an essential milestone for robotics and an area of active research with fundamental work needed ahead \cite{billard2019trends}.
However, the human manipulation skill that has attracted the most attention from roboticists is prehensile manipulation, or \textsl{grasping}.
Manipulation by grasping is attractive primarily because, once an object is grasped, it generally does not need to be tracked over time and uncertainty on its state is reduced. However, grasping is limited in capability by i) reachability of the robot arm, ii) mechanical design limitations of the end-effector, iii) physical properties of the object being manipulated, and iv) accuracy of the perception system.

\textsl{Non-prehensile manipulation} (i.e., any kind of manipulation not involving grasping, hereinafter referred to as NPM) offers a complementary solution to prehensile manipulation by significantly expanding the size (intended as the set of reachable configurations) and dimensionality (intended as the number of degrees of freedom) of the operational space of even the simplest robot manipulator \cite{lynch1999dynamic}.
In other words, NPM can be used to manipulate objects when conventional grasping-based manipulation is infeasible or unnecessary.
Realistic robot applications might expect the robot to operate in dense clutter, in the presence of occlusions, or in ungraspable configurations---for example, the target object is in a pose that is not directly reachable by the end-effector or the target object is too large or too heavy. These applications may result in failure modes for robot operation through traditional grasping. Consequently, in such situations it is beneficial to complement the robot's skillset with NPM primitives. Indeed, NPM can be used both as a skill and a failure recovery mechanism, which points to the versatility of this paradigm.

\begin{figure}
  \centering
    \includegraphics[width=.9\linewidth]{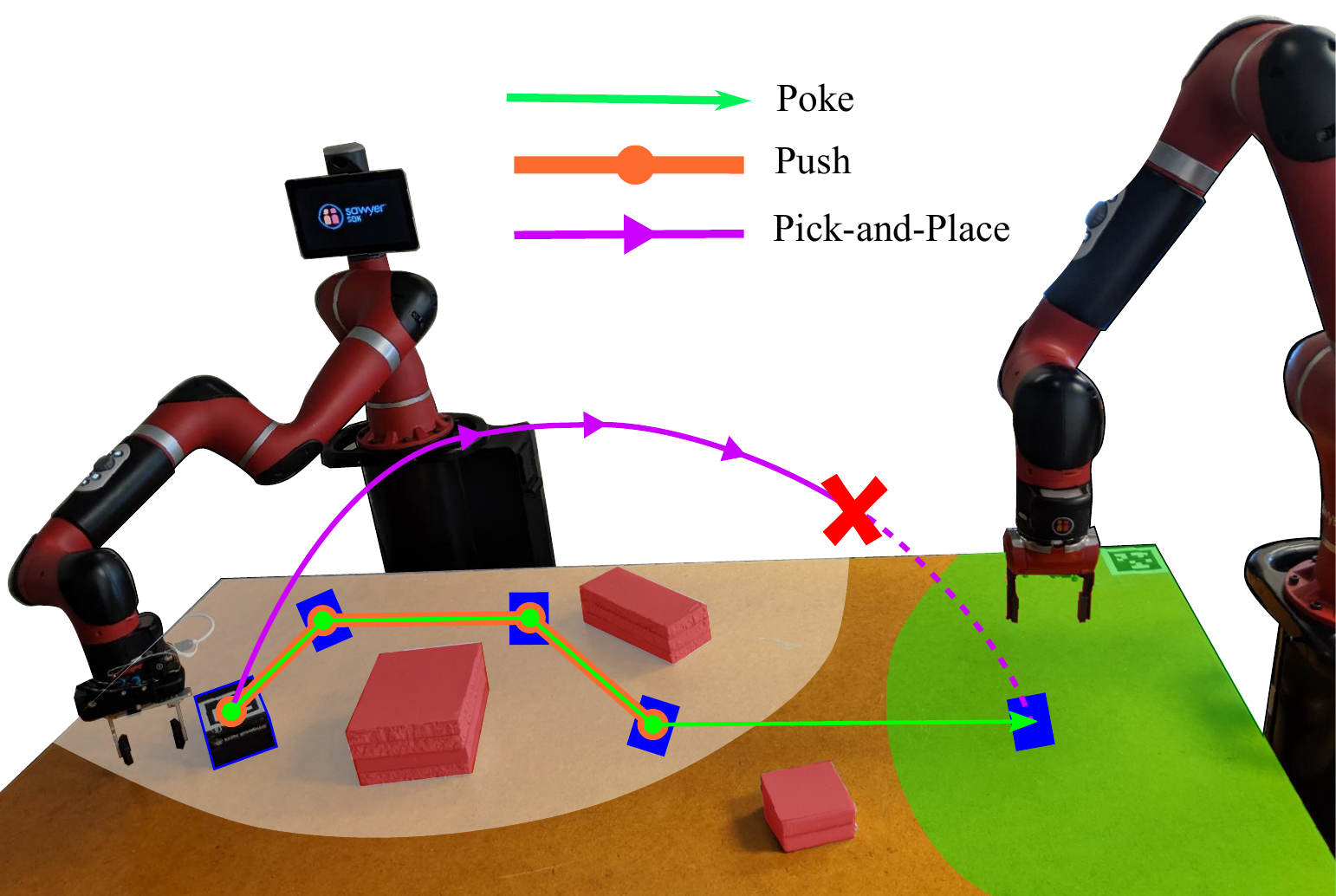}
  \vspace{-4pt}
  \caption{This work demonstrates poking as a skill and a failure recovery tactic to increase the portfolio of capabilities at the robot's disposal. Here, an object (blue) is located in an obstacle-rich (red) workspace with non-overlapping reachable regions for each robot defined by beige and green shading. The first robot manipulates the object into the green region successfully via poking (path shown in green), but fails to do so via pushing or grasping (paths shown in orange and purple, respectively).\vspace{-14pt}}\label{fig:firstpg}
\end{figure}

In this work, we demonstrate the utility of NPM through poking, a skill that allows fast object manipulation and expands the size of a manipulator's reachable workspace. The basic idea behind this work is detailed in \cref{fig:firstpg}.
\textsl{Poking} is a NPM primitive wherein a robot end-effector applies an instantaneous force to an object of interest to set the object in planar translational and rotational motion (\textsl{impact} phase). The object eventually slows down and comes to rest due to Coulomb friction (\textsl{free-sliding} phase). Poking has a multitude of desirable properties that makes it complementary to grasping and serves as a generalized form of pushing where applied impulse forces are low \cite{Huang-1997-14452, pasricha2021pokerrt}. 
These characteristics make poking especially suitable for industrial settings where robots need to operate in their delineated workspaces alongside other robots or humans while still being able to pass objects. Additionally, in logistics or e-commerce settings, poking can be used for either stowing objects in boxes or in synergy with pick-and-place to optimize operations by increasing speed and coverage area.
In this paper, we design and implement two closed-loop, kinodynamic, sampling-based planners called \textsl{PokeRRT} and \textsl{PokeRRT*} which decouple skill modeling and path planning and specifically focus on leveraging the following advantages of poking over pushing and grasping:
i) it does not require constant contact between the manipulator and the object, therefore greatly expanding the size of the manipulator's workspace,
ii) it does not impose restrictions on the shape or size of objects that can be manipulated, and
iii) it is inherently faster and therefore capable of covering large distances in short periods of time.

After discussing related work (\cref{sec:background}), we present \textsl{PokeRRT} which leverages a simulation engine to generate a collision-free path between two points in the object configuration space (\cref{sec:methods}). We conclude with an experimental validation (\cref{sec:evaluation}) and discussion (\cref{sec:discussion}) of \textsl{PokeRRT} in both simulated and real-world settings.%

\section{Background and Related Work}\label{sec:background}

Research in non-prehensile manipulation dates back to the 90s \cite{lynch1999dynamic, ruggiero2018nonprehensile, toussaint2018differentiable}, with the vast majority of prior work leveraging heuristics or analytical models based on simplifying assumptions. Related work has focused on skills such as throwing \cite{zeng2020tossingbot}, poking \cite{Huang-1997-14452}, and pushing \cite{woodruff2017planning,RSS2020Changkyu,mason1986mechanics}.
More recently, pushing manipulation has received increased attention due to availability of large-scale datasets and the inherent controllability of the skill \cite{bauza2019omnipush}. Pushing operates under the quasistatic assumption (i.e., the inertial effects of robot--object and object--environment interaction are ignored) to reduce modeling complexity, thereby limiting robot velocities and accelerations. On the contrary, poking must plan around the non-negligible effects of inertial forces: the object continues moving after robot--object contact is broken, thus allowing for faster planar manipulation of objects. Past work in pushing manipulation incorporates simulation and analytical models in a motion planning loop to get the next feasible state \cite{zito2012two}\cite{zhou2019pushing}. Additional contributions in push modeling involve combining object state estimation with affordance prediction from image data to determine contact points for achieving the optimal push \cite{kloss2020accurate} and creating a deep recurrent neural network model to model push outcomes for a variety of objects \cite{li2018push}. However, both approaches use a greedy planner operating in obstacle-free environments. In this work, we propose a sampling-based kinodynamic framework that is capable of planning dynamic collision-free paths in the object configuration space. 

Past work on sampling-based planning techniques for non-prehensile manipulation involve kinodynamic approaches to rearrangement planning and whole-arm manipulation. This work is constrained  by the quasistatic assumption, thereby limiting object manipulation speed through the use of non-dynamic primitives \cite{king2015nonprehensile}. However, when operating in a dynamic regime, lack of kinematic modeling of the robot and lack of contact modeling between the robot arm and environmental objects may result in unusable actions explored during the planning phase \cite{haustein2015kinodynamic}. These approaches also employ an open-loop paradigm and do not take robot, sensing, and model uncertainty into account. Moreover, open-loop kinodynamic planners that consider uncertainty in the planning process may generate conservative plans that do not exploit robot and object dynamics to their full extent \cite{8207585}. Our approach uses a simulation engine that considers robot kinematics and contact while planning and operates in a closed-loop manner, replanning if the resultant pose is outside a certain threshold.
% \ap{Additionally, a key difference to prior work is the explicit consideration of manipulator constraints and the extension \textsl{PokeRRT*} that is used to produce shorter solutions to leverage poking's crucial ability to cover large distances quickly.}
% Additionally, neither approach leverages RRT*, thereby not exploiting the long action paths that are crucial for taking full advantage of dynamic primitive-based planning. -- but here is work on kinodynamic RRT* so probably shouldn't mention this

In all, evidence from prior work suggests that \textsl{poking}---sometimes referred to as ``releasing'' or ``impulsive manipulation''---is a relatively unexplored primitive. Analytical models for poking dynamics are restricted to rotationally symmetric objects or situations where pusher--object contact can be geometrically modeled \cite{Huang-1997-14452, zhu2005frictional}. The need for specialized impulse-delivery apparatus to achieve poking is explored in \cite{Huang-1997-14452}. This makes planar manipulation of objects cumbersome due to the manual relocation of the apparatus required. In this work, we use a standard open-chain robot arm equipped with an electric parallel gripper to deliver impulses using joint space velocity control.

In order to generate impulse-based action paths that obey the robot's kinematic constraints and produce feasible object motion, we use a PyBullet \cite{coumans2019} simulation environment as the forward physics model in the planning loop.
% Since methods for skill modeling are inextricably tied to the research on NPM planning detailed above, as a dynamical model is a prerequisite of any NPM system.
% Specifically, we use PyBullet \cite{coumans2019} as the forward physics model in our planner. 
While simulation may not accurately capture real-world dynamics \cite{collins2019quantifying}, the closed-loop nature of our planning framework compensates for this approach. This effective use of simulation captures essential characteristics of robot and object dynamics at faster than realtime speeds and in a safe manner without inducing significant wear-and-tear caused by dynamic trajectory execution on a real robot.

\section{Methods}\label{sec:methods}

In this section, we present an overview of the proposed approach for poking (\cref{sec:poking}), its characterization in a simulated environment (\cref{sec:sim-model}), and our motion planning algorithms, \textsl{PokeRRT} and \textsl{PokeRRT*}, that leverages the simulated environment to plan a collision-free path for the target object through its configuration space (\cref{sec:poke-rrt}). We also introduce empirically-driven heuristics to \textsl{PokeRRT} that leverage the large and quick displacement property intrinsic to poking manipulation.\vspace{-4pt}

\begin{figure*}
\vspace{4pt}
\centering
\includegraphics[width=\linewidth]{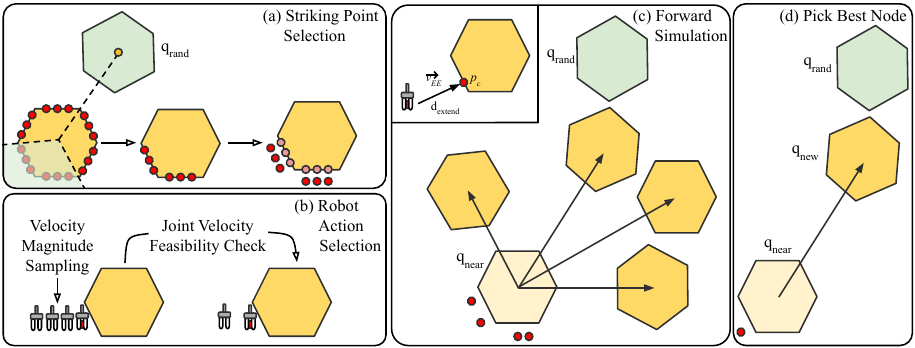}
\caption{Path planning for poking consists of 2 steps: action sampling (a, b) and graph expansion (c, d). (a) Points are sampled uniformly on the object contour (red) and filtered through a conical region originating from the target position (green). Striking points are generated by extending away from contour points in the normal direction. (b) End-effector velocity magnitudes are sampled for each striking point and filtered out if joint velocities are infeasible due to mechanical limitations of the robot. (c) Feasible actions are applied in simulation to get resultant poses. (d) The resultant pose closest to the target position is added to the planning graph.\vspace{-12pt}}\label{fig:blockdiag}
\end{figure*}

\subsection{Formalization of the Poking Motion Primitive}\label{sec:poking}

Poking manipulation is modeled as a process composed of two phases: i) \textsl{impact}, where the robot end-effector makes instantaneous contact with the object; and ii) \textsl{free-sliding}, where the object slides on a planar surface and comes to a stop due to Coulomb friction.
As detailed in \cref{fig:blockdiag}c, two parameters are required to describe the first phase of poking: the point of contact $p_c$ (i.e., where on the contour of the object to strike), and the magnitude of the impact velocity $\|\vec{v}_{EE}\|$.
Slippage at the contact point between the end-effector and the object may lead to non-linearities; therefore, we fix the direction of $\vec{v}_{EE}$ as being normal to the object's contour.
Importantly, in order to apply an instantaneous force, the robot must come to a complete halt upon contact with the object. Therefore, collision between the end-effector and the object is treated as an elastic collision. The motor torques applied to stop the end-effector upon contact prevent the impulsive interaction from being truly elastic; however, this can be safely ignored by stopping the end-effector slightly past the contact point.

Given a uniform-density object of mass $m$ with a coefficient of friction $\mu$ whose center of mass starts in an initial planar pose $q_i = (x_i, y_i, \theta_i)$  and the control parameters (see \cref{fig:blockdiag}c) that determine the contour of the object to strike ($p_c$) and the velocity magnitude ($\|\vec{v}_{EE}\|$), we can solve the second phase of poking and determine the final pose of the object $q_f = (x_f, y_f, \theta_f)$ using a physics simulation engine.

\subsection{Simulation Model for Poking}\label{sec:sim-model}

In order to understand the effects of impulsive forces on objects, we use the PyBullet physics simulation engine \cite{coumans2019}. PyBullet models rigid body dynamics by performing numerical integration over time with equations of motion to solve for object position and velocity. Joint constraints, contact forces, and friction, in addition to external forces such as gravity, are taken into account in the forward dynamics solver of the engine and modeled as constraints in a Linear Complementarity Problem (LCP).

The main point of interest when modeling impulsive interactions is contact force. In the context of poking,  contact is primarily dominated by frictional interactions between i) the robot end-effector and the object of interest, and ii) the object and the environment (i.e., the object's planar support surface and surrounding obstacles). Analytical models for contact make simplifying assumptions that do not fully represent the complexity of real-world dynamics, including the nonlinear nature of friction and actuator degradation and latency. Additionally, they do not generalize well to a diverse set of objects.
Conversely, learning-based approaches, while capable of achieving generalization and modeling uncertainty and complexity, are data-inefficient. Collecting abundant and task-representative data in real-world robotics is expensive with regard to the time taken and the wear-and-tear caused on a real robot through repetitive, and potentially, high-acceleration trajectory executions such as those characteristic to poking manipulation. Executing such trajectories and modeling collisions along the action path with the object and the environment is therefore far safer in simulation.

While simulation does not perfectly capture the inherent complexity and stochasticity of real-world contact dynamics due to the simplistic nature of the underlying analytical models used, it nonetheless provides a good balance between pure learning and analytical models by ensuring interaction modeling is both cheap and safe while encapsulating the essential characteristics of robot--object and object--environment interactions. The closed-loop nature of our proposed motion planner also compensates for any inaccuracies in simulation modeling while executing poke plans in the real-world. That is, if a resultant pose violates a predefined object pose threshold, we compute a new poke plan from the current object pose to the goal region. This crucial feature allows our planner to plan feasible poke paths in simulation and execute them in the real-world.\vspace{-4pt}

\subsection{Motion Planning for Poking}\label{sec:poke-rrt}

Using simulation as the forward physics model in our planning loop, we design and implement \textsl{PokeRRT} that takes advantage of the inherent speed and efficiency of poking manipulation.
This global path planning approach leverages goal and obstacle information in object configuration space to introduce a bias into motion planning and to keep the sampling space low-dimensional to ensure fast planning.

\subsubsection{Object Configuration Space}

Our planner operates in $\mathcal{C}$, the continuous ($x$, $y$, $\theta$) configuration space of the object. $x$ and $y$ are confined by the limits of the planar workspace, whereas $\theta$ is defined in the range $[-180^\circ, 180^\circ)$.

\subsubsection{Action Sampling}\label{sec:action-sampling}

The proposed planner decouples skill modeling and object path planning to allow for the evaluation of multiple skill models which directly improve planning outcomes. We achieve this through an action-oriented approach by employing a series of filtering steps to pick a set of valid actions $\boldsymbol{a_{valid}} = \langle \boldsymbol{p_c}, \boldsymbol{\|\vec{v}_{EE}\|}\rangle$ to apply at a given object configuration $q$. A sampling approach that yields actions which can be simulated in a physics engine is desirable since inverse modeling of frictional contact and object motion may either be intractable or not guarantee feasible robot motion.

To achieve the correct direction of motion, candidates for object contour points that are ideal for impact must lie on the side nonadjacent to the target object position. Given an object configuration $q$, contour points $\boldsymbol{p_{contour}}$ are chosen as acceptable candidates if they lie within a cone originating from a target pose $q_{rand}$ (\cref{fig:blockdiag}a). The target pose is sampled during path planning as described in \cref{sec:planners}.
For each candidate contour point $p_c \in \boldsymbol{p_{contour}}$, a striking point $p_s$ is computed at a fixed distance, $d_{extend}$, from $p_c$ in the normal direction away from the object (\cref{fig:blockdiag}c). The robot engages in joint space velocity control to apply a poke in its operational space by moving from $p_s$ to $p_c$ at velocity $\vec{v}_{EE}$. 
Collision-free robot joint configurations, $\boldsymbol{\theta}_s$ and $\boldsymbol{\theta}_c$, for both $p_s$ and $p_c$ are computed using \textsl{TRAC-IK} \cite{7363472}. Since impactful contact is a core requirement for successful manipulation, we do not check for collisions between the robot end-effector and the object.
A final filter is applied to see if $\boldsymbol{\dot{\theta}}_c = J^{-1}(\boldsymbol{\theta}_c)\vec{v}_{EE}$, i.e., the joint velocities at $\boldsymbol{\theta}_c$ given the given operational space velocity $\vec{v}_{EE}$, satisfies the joint velocity thresholds for the robot, $\langle\boldsymbol{\dot{\theta}}_{min}$, $\boldsymbol{\dot{\theta}}_{max}\rangle$ (\cref{fig:blockdiag}b). 

The kinematic feasibility check along the entire path from $\boldsymbol{\theta}_s$ to $\boldsymbol{\theta}_c$ is provided by the simulation engine, i.e., the success or failure of executing the sampled action in simulation will determine whether our planner adds this action to the planning graph.
Given this process for generating feasible actions that move the object from its current position towards the target position, we build a global path planner to generate an action path through object configuration space from the object's current pose to task goal region.

\subsubsection{Path Planning}\label{sec:planners}

In this section, we introduce a novel motion planner named \textsl{PokeRRT}; it generates a graph of feasible robot actions in the object configuration space.
With the aforementioned $\boldsymbol{a_{valid}}$ as a set of valid control parameters, exploration properties of the rapidly-exploring random tree (RRT) algorithm \cite{lavalle1998rapidly} are leveraged to generate a planning graph. The integration of poking-specific control inputs to RRT ensures that all nodes in the planning graph are achievable configurations, while the RRT algorithm ensures exploration bias to the largest Voronoi regions of the configuration space. This bias is especially central to poking manipulation due to its inherent ability to cover large distances in the operational space.

A new node $q_{rand}$ is set as the goal node with probability $p_{bias}$, otherwise $q_{rand}$ is randomly sampled in the object configuration space. Actions from $\boldsymbol{a_{valid}}$ are applied from the nearest graph node $q_{near}$ towards $q_{rand}$ (\cref{fig:blockdiag}c) and the resultant node $q_{new}$ that minimizes the distance to $q_{rand}$ is added to the graph (\cref{fig:blockdiag}d; GET\_BEST\_RESULT() in \cref{algo:pokerrt*}).
Resultant poses are added to the planning tree until the task goal region $\mathcal{Q}_{goal}$ is reached, at which point the path from $q_{start}$ to $\mathcal{Q}_{goal}$ in the planning tree is extracted by backtracking (\cref{algo:pokerrt*}, Line 13). \textsl{PokeRRT} does not find the absolute least-cost path since RRT is not asymptotically optimal by nature \cite{lavalle1998rapidly}. The current implementation operates in a closed-loop paradigm---we replan on the fly if the resultant pose after a poking action is outside a specified threshold to compensate for inaccuracies of the simulation poking model and collisions between the object and environmental obstacles. The replanning threshold for an object is decided based on a collection of pokes executed in simulation and the real-world, i.e. the average object resultant pose error caused by the \textsl{sim2real} gap serves as an empirical estimate of the threshold. Therefore, this measure encompasses any uncertainty resulting from sensor, actuation, and skill model noise. The replanning strategy is particularly important for poking since frictional interactions between the object and the support surface are stochastic in nature and motor slippage and nonlinearities in joint modeling in the real-world can induce noise in robot--object interaction. Additionally, unlike pushing and grasping, poking lacks full controllability which induces further uncertainty on the final object pose.
Using this planner, poking can be used both as a skill and as a failure recovery tactic and can be shown to operate faster and in a more diverse set of scenarios than pushing or grasping.
Since poking is inherently capable of covering larger distances in short periods of time due to high impact interactions, we can leverage this insight and add extensions to \textsl{PokeRRT} to plan sparse paths in object configuration space.\vspace{-8pt}

\begin{algorithm}\footnotesize
\caption{PokeRRT*}\label{algo:pokerrt*}
  \KwIn{Start Node $q_{start}$, Goal Region $\mathcal{Q}_{goal}$, Neighborhood Radius $r$, Object Configuration  Space $\mathcal{C}$, Poke Dataset $\mathcal{D}$}
  \KwOut{Poke Path $\mathcal{P}$}
  $\mathcal{G}.$add\_vertex($q_{start}$);\\
  \While{not GOAL\_REACHED($\mathcal{G}, \mathcal{Q}_{goal}$)} {
      $q_{rand} =$ RAND($\mathcal{C}$); $q_{near} =$ NEAREST($q_{rand}, \mathcal{G}$);\\
      $q_{new}' = q_{near} + $ SAMPLE\_DISP($\mathcal{D}, q_{rand}$);\\
      $q_{min} = \underset{N\_POKES}{\arg\min}$ RADIAL\_NN($q_{new}', \mathcal{G}, r$);\\
      \vspace{4pt}
    %   $q_{min} =$ CHOOSE\_PARENT($q_{new}', \mathcal{G}, r$);\\
      $q_{new} =$ GET\_BEST\_RESULT($q_{min}, q_{new}', \mathcal{G}$);\\
      $\mathcal{G}.$add\_vertex($q_{new}$); $\mathcal{G}.$add\_edge($q_{min}, q_{new}$);\\
    %   \tcp{rewiring, 9-14}
    %   $\boldsymbol{q_{neighbors}} =$ RADIAL\_NN($q_{new}, \mathcal{G}, r$);\\
      \ForEach{$q \in$ RADIAL\_NN($q_{new}, \mathcal{G}, r$)}{
        \If{N\_POKES($q_{new}$) $+ 1 <$ N\_POKES($q$) and IS\_LEAF\_NODE($q$)}{
            $\mathcal{G}.$remove\_edge(PARENT($q$), $q$);\\
            $q_{result} =$ GET\_BEST\_RESULT($q_{new}, q, \mathcal{G}$);\\
            $\mathcal{G}.$add\_edge($q_{new}, q_{result}$);\\
        }
      }
  }
$\mathcal{P} = $ BACKTRACK($\mathcal{G}$, $q_{start}$, $\mathcal{Q}_{goal}$);\\
\textbf{return} $\mathcal{P}$ 
\end{algorithm}
% \vspace{-20pt}
\vspace{-10pt}

\begin{figure}
\vspace{6pt}
  \begin{minipage}[b]{\linewidth}
  \centering
  \resizebox{0.8\textwidth}{!}{
    
\begin{tabular}{|c|c|c|c|c|c|}
\hline
\textbf{} &
  \textbf{Scenario Description} &
  \textbf{Poke} &
  \textbf{Push} &
  \textbf{Grasp}\\ \hline
\textbf{S1} & No Obstacles & \checkmark & \checkmark & \checkmark \\ \hline
\textbf{S2} & 2 Obstacles  & \checkmark & \checkmark & \checkmark \\ \hline
\textbf{S3} & 4 Obstacles  & \checkmark & \checkmark & \checkmark \\ \hline
\textbf{S4} & Wide Object  & \checkmark & \checkmark & \\ \hline
\textbf{S5} & Tunnel       & \checkmark &  & \checkmark \\ \hline
\textbf{S6} & Shared Workspace & \checkmark & & \\ \hline
\end{tabular}
}
\end{minipage}\vskip 4pt
\centering
\includegraphics[width=\linewidth]{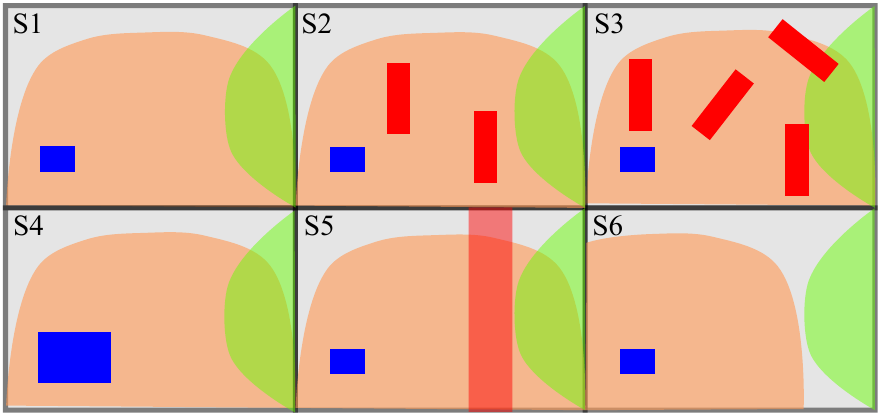}
\caption{The robot successfully pokes the object (blue) from its reachable workspace (orange) to the goal region (green) in all scenarios while avoiding obstacles (red). \textsl{PokeRRT}, \textsl{PokeRRT*}, and baseline algorithms are evaluated in 6 scenarios---no obstacles (S1), $2$ obstacles (S2), $4$ obstacles (S3), wide object (S4), tunnel (S5), and non-overlapping shared workspace (S6). The robot is unable to i) push or pick-and-place in S6 due to limited robot reach, ii) push in S5 due to workspace obstruction in the action path, and iii) pick-and-place in S4 due to object being wider than gripper width.\vspace{-12pt}}\label{fig:scenarios}
\end{figure}

\subsubsection{Online Path Smoothing}

To minimize the number of pokes and enable poke planning to cover large distances, we combine empirically-derived heuristics with insights from RRT* \cite{karaman2011sampling} to develop \textsl{PokeRRT*} (\cref{algo:pokerrt*}).
RRT* allows for the discovery of new lower-cost paths until the goal region is reached by introducing two additional steps to RRT---choose-parent and rewiring.
In the context of \textsl{PokeRRT*}, choosing the best parent node $q_{min}$ replaces the edge from $q_{near}$ to $q_{new}$ with the edge from $q_{min}$ to $q_{new}$. Planning time for \textsl{PokeRRT*} can be reduced by using a data-driven heuristic to generate $q_{new}'$---instead of applying actions from $\boldsymbol{a_{valid}}$ to get $q_{new}'$, we sample $\Delta q$ from a range of displacements from a dataset of random pokes $\mathcal{D}$.
$q_{new}'$ is then computed as $q_{near} + \Delta q$ in the direction of $q_{rand}$ (\cref{algo:pokerrt*}, Line 4).

Then $q_{min}$ is chosen as the neighbor within radius $r$ of $q_{new}'$ that minimizes the number of pokes from $q_{start}$ to $q_{new}'$ (\cref{algo:pokerrt*}, Line 5). Radius $r$ is selected empirically as the average displacement of the object in the dataset $\mathcal{P}$.
The best action from $q_{min}$ to $q_{new}'$ is recomputed by sampling actions from $\boldsymbol{a_{valid}}$ to make sure each edge in the graph is a valid action. Resultant pose $q_{new}$ is chosen as one that minimizes the distance to $q_{new}'$. $q_{new}$ and the edge from $q_{min}$ to $q_{new}$ are added to the graph.
In the rewiring step, the neighborhood of $q_{new}$ is rewired to minimize the number of pokes along the path from $q_{start}$ through $q_{new}$ to a neighboring leaf node (\cref{algo:pokerrt*}, Lines 8-12). Note that \textsl{PokeRRT*} does not use the extension to RRT* that exploits the anytime nature of RRT* to continuously improve path quality by allocating planning time \cite{karaman2011anytime}---choose-parent and rewiring are the primary modes for online path sparsification.
As a result, \textsl{PokeRRT*} leads to fewer pokes in the planned path than \textsl{PokeRRT}, thereby taking advantage of poking's core competency of allowing larger displacements.

\section{Evaluation}\label{sec:evaluation}\vspace{-2pt}

In this section, we qualitatively and quantitatively evaluate \textsl{PokeRRT} and \textsl{PokeRRT*} in simulation and the real-world under various environment setups.\footnote{Links to our code, videos, and demonstrations of the experiments are available here: \href{https://hiro-group.ronc.one/poke-rrt-icra-22.html}{https://hiro-group.ronc.one/poke-rrt-icra-22.html}.} We measure success rates, task times (in seconds), and number of executed actions in the final planned path for both motion planners. Success rates are averaged across all scenarios for a given planner, whereas task times and number of executed actions are presented separately for each scenario. Task time is defined as the sum of planning, execution, and replanning times. Object start pose is kept fixed across various trials to aid reproducibility of results. Simulation and real-world results are averaged over 250 and 10 trials, respectively.
Collectively, our results demonstrate that \textsl{PokeRRT} and \textsl{PokeRRT*} indeed enable fast and successful manipulation of objects in conditions where push planning and pick-and-place fail.

\subsection{Experimental Setup}

\begin{table*}[]
\vspace{4pt}
\resizebox{\textwidth}{!}{
\centering
\begin{tabular}{c|c|c|c|c|c|c|c|}
\cline{2-8}
 &
  \multicolumn{6}{c|}{\textbf{Task Time [seconds]}}&\multicolumn{1}{c|}{\textbf{Success Rate}}\\ \hline
\multicolumn{1}{|c|}{\textbf{Planner}} &
  \textbf{S1} &
  \textbf{S2} &
  \textbf{S3} &
  \textbf{S4} &
  \textbf{S5} &
  \textbf{S6} 
  & \textbf{S1 - S6}\\ \hline
\multicolumn{1}{|c|}{PokeRRT*} & 
\textbf{46.43 (21.67)} & 
212.74 (100.01) & 
232.42 (118.63) & 
70.47 (38.82) & 
132.62 (98.27) & 
\textbf{130.01 (84.05)} & 
0.87 (0.31) \\ \hline
\multicolumn{1}{|c|}{PokeRRT} & 
49.69 (21.11) & 
\textbf{196.54 (124.61)} & 
\textbf{167.55 (138.63)} & 
\textbf{64.14 (52.45)} & 
\textbf{116.35 (89.72)}& 
171.77 (103.21) & 
\textbf{0.88 (0.31)} \\ \hline
\multicolumn{1}{|c|}{\begin{tabular}[c]{@{}c@{}}LI-PokeRRT*\end{tabular}} & 
169.87 (71.28) & 
378.90 (122.21) & 
346.70 (111.71) & 
165.91 (44.87) & 
N/A & 
N/A & 
0.53 (0.28) \\ \hline
\multicolumn{1}{|c|}{\begin{tabular}[c]{@{}c@{}}LI-PokeRRT\end{tabular}} & 
123.71 (36.72) & 
328.66 (107.18) & 
322.28 (138.05) &
135.80 (32.91) & 
N/A & 
N/A & 
0.63 (0.18) \\ \hline
\multicolumn{1}{|c|}{\begin{tabular}[c]{@{}c@{}}Push Planner\end{tabular}} & 
122.68 (47.07) & 
284.78 (104.40) & 
249.32 (68.30) & 
117.58 (67.05) & 
N/A & 
N/A & 
0.44 (0.26) \\ \hline\hline
\multicolumn{1}{|c|}{Pick-and-Place} & 
16.29 (3.76) & 
17.71 (3.81) & 
16.14 (3.25) & 
N/A & 
19.15 (4.99) & 
N/A & 
0.67 (0.00)\\ \hline
\end{tabular}
}
\caption{Task times [\textsl{mean (stddev)}] for various planning algorithms in simulation are presented. Success rates are aggregated across all 6 scenarios. Overall, poke planning is faster than push planning and leads to higher success rate than pushing or grasping. Low-impulse (LI) poke planners and the push planner are unsuccessful in S5 and S6 due to action path collisions (S5) or goal region being outside the robot's reachable workspace (S6). The robot is unable to pick or place object in S4 and S6  due to limitations in robot mechanics.\vspace{-12pt}}
\label{tab:eval-sim-time}
\end{table*}

\subsubsection{Scenarios}\label{sec:evaluation-scenarios}

We test multiple algorithms for poking, pushing, and grasping manipulation in six scenarios (\cref{fig:scenarios}).
% \ap{\textsl{PokeRRT} and \textsl{PokeRRT*} expand a robot's reachable workspace, in contrast to pushing and pick-and-place which operate within this region. Pushing also operates under the quasistatic assumption, as a result of which the robot must maintain constant contact with the object. And finally, pick-and-place is also limited by the object dimensions, i.e., if the object is bigger than the gripper width, pick-and-place will fail.}
Our motivating application is the concurrent operation of two robots with potentially non-overlapping reachable regions located in adjacent workcells on a factory floor. Therefore, the objects being manipulated must be accessible by both robots to enable successful collaboration.
Given this motivation, scenarios are designed to test the flexibility of the planner in generating plans for manipulating the object in a planar workspace from a fixed start pose to a goal region, which represents the workspace of a second robot. To ensure that object--obstacle collisions are handled robustly through the replanning strategy, we enforce the following properties in our scenarios: i) the objects used in each scenario are rigid and therefore retain their shape across multiple trials and ii) obstacles are fixed to the table so that collisions do not change the planning configuration space.

Scenarios 1-3 (S1-S3 in \cref{fig:scenarios}) are designed to test baseline manipulation capability through uncluttered (S1) and cluttered (S2, S3) environments. S2 and S3 contain 2 and 4 obstacles, respectively, at fixed poses in the shared robot workspace. The object being manipulated has size $[9, 14, 5]$ cm and mass $m=87$ g. Poke planning, push planning, and pick-and-place will all work in these scenarios since the goal region overlaps with the first robot's reachable workspace.
Scenario 4 (S4 in \cref{fig:scenarios}) contains a bigger cuboid object of size $[11, 18, 11]$ cm and mass $m=112$ g in an obstacle-free workspace. Poke planning and push planning will work in this scenario, however pick-and-place will not since gripper width ($10$ cm) is smaller than the minimum dimension of the cuboid. This scenario represents situations where object properties are incompatible with robot kinematics, i.e., if the object of interest is too heavy or too wide to be grasped or if the end-effector is malfunctional, and the robot needs to formulate an alternate manipulation plan.

Scenario 5 (S5 in \cref{fig:scenarios}) presents two work cells with a tunnel in the center of the table. The first robot cannot push the object to the second robot because the end-effector will collide with the divider along its action path. However, poke planning will succeed due to the short duration of robot--object contact. The first robot is able to grasp the object over the divider in our setup but for larger dividers such as a screen, pick-and-place will fail.
Scenario 6 (S6 in \cref{fig:scenarios}) contains an obstacle-free workspace with non-overlapping reachable regions for each robot. The first robot is able to poke the target object to the second robot's reachable workspace, but cannot push or pick-and-place due to limited reachability. Since pushing operates under the quasitstatic assumption, it requires constant contact between the robot end-effector and the object being manipulated and therefore, manipulation is limited by robot kinematics and reachability. This scenario represents cases where task success is limited by the robot's kinematic characteristics, i.e., if the goal pose is outside the robot's reachable workspace.

\subsubsection{Parameters}

The action vector $\boldsymbol{a_{valid}} = \langle \boldsymbol{p_c}, \boldsymbol{\|\vec{v}_{EE}\|}\rangle$ for our proposed motion planners is generated at runtime. Contour points $p_c$ close to object corners are ignored for clean pokes. The velocity magnitudes $\|\vec{v}_{EE}\|$ are sampled in the $[0.3, 1.0]$ m/s range to increase likelihood of robot joints achieving the commanded operational space velocities. Goal region is defined as the workspace of the second robot, determined as the frequency of inverse kinematics poses achievable on discretized xy-locations on the planar support surface of the object. The empirically-determined replanning threshold is $5cm$ and $10^{\circ}$.

\subsubsection{Algorithms}

\textsl{PokeRRT} and \textsl{PokeRRT*} are compared against several baseline approaches. To evaluate pushing, we use the \textsl{Two-Level Push Planner} presented in \cite{zito2012two} since it also uses simulation in the planning loop to get the next feasible environment state. This work integrates simulation-based forward modeling with sampling-based motion planning to explore the space of feasible pushing actions required to get an object from start to goal. Our baseline approaches, \textsl{Low-Impulse (LI) PokeRRT} and \textsl{Low-Impulse (LI) PokeRRT*}, are designed to show that poking is a more fundamental manipulation skill and encompasses pushing if the applied impulse magnitudes are kept small. They operate similarly to \textsl{PokeRRT} and \textsl{PokeRRT*} but with $\|\vec{v}_{EE}\| = 0.2$ m/s to simulate pushing. Any replanning for a given planner is done using that same planner. Lastly, \textsl{Pick-and-Place} is performed in an open-loop manner with predefined grasps for known objects. An experimental trial fails if the planner does not find a valid plan to the goal region in 240 seconds or if the object falls off the table during execution.\vspace{-4pt}

\begin{table}[]
\resizebox{\columnwidth}{!}{
\centering
\begin{tabular}{c|c|c|c|c|c|c|}
\cline{2-7}
                                                                                     & \multicolumn{6}{c|}{\textbf{Number of Actions in Execution Path}}                                 \\ \hline
\multicolumn{1}{|c|}{\textbf{Planner}}                                               & \textbf{S1} & \textbf{S2} & \textbf{S3} & \textbf{S4} & \textbf{S5} & \textbf{S6} \\ \hline
\multicolumn{1}{|c|}{PokeRRT*}                                                       & 
\textbf{\begin{tabular}[c]{@{}c@{}}3.93\\ (0.98)\end{tabular}} & \textbf{\begin{tabular}[c]{@{}c@{}}4.62\\ (1.04)\end{tabular}}                       
&   \textbf{\begin{tabular}[c]{@{}c@{}}4.49\\ (1.02)\end{tabular}}   
&   \textbf{\begin{tabular}[c]{@{}c@{}}4.76\\ (1.20)\end{tabular}}   
& \textbf{\begin{tabular}[c]{@{}c@{}}3.71\\ (1.02)\end{tabular}} &  \textbf{\begin{tabular}[c]{@{}c@{}}3.65\\ (1.08)\end{tabular}}\\ \hline
\multicolumn{1}{|c|}{PokeRRT} &
  \begin{tabular}[c]{@{}c@{}}6.10 \\ (1.96)\end{tabular} &
\begin{tabular}[c]{@{}c@{}}8.21\\ (2.95)\end{tabular}  &
\begin{tabular}[c]{@{}c@{}}7.94\\ (2.08)\end{tabular}  &\begin{tabular}[c]{@{}c@{}}6.77\\ (2.35)\end{tabular}
  &
\begin{tabular}[c]{@{}c@{}}4.56\\ (1.27)\end{tabular}  &
 \begin{tabular}[c]{@{}c@{}}8.17\\ (2.46)\end{tabular} \\ \hline
\multicolumn{1}{|c|}{\begin{tabular}[c]{@{}c@{}}LI-PokeRRT*\end{tabular}} & 
\begin{tabular}[c]{@{}c@{}}16.43\\ (2.28)\end{tabular}&
\begin{tabular}[c]{@{}c@{}}22.27\\ (2.26)\end{tabular}
&\begin{tabular}[c]{@{}c@{}}19.00\\ (1.54)\end{tabular}
&\begin{tabular}[c]{@{}c@{}}18.54\\ (1.44)\end{tabular}
&N/A
& N/A            \\ \hline

\multicolumn{1}{|c|}{\begin{tabular}[c]{@{}c@{}}LI-PokeRRT\end{tabular}} &
\begin{tabular}[c]{@{}c@{}}19.08 \\ (1.51)\end{tabular}
  &
\begin{tabular}[c]{@{}c@{}}19.62\\ (1.64)\end{tabular}  &
 \begin{tabular}[c]{@{}c@{}}20.74\\ (2.13)\end{tabular} &\begin{tabular}[c]{@{}c@{}}19.38\\ (1.59)\end{tabular}
  &
N/A  &
N/A  \\ \hline
\multicolumn{1}{|c|}{\begin{tabular}[c]{@{}c@{}}Push Planner\end{tabular}} &
\begin{tabular}[c]{@{}c@{}}8.72 \\ (1.42)\end{tabular} &
\begin{tabular}[c]{@{}c@{}}8.61\\ (1.71)\end{tabular}   &
\begin{tabular}[c]{@{}c@{}}7.68\\ (1.19)\end{tabular}   &\begin{tabular}[c]{@{}c@{}}8.53 \\ (1.51)\end{tabular}
  &N/A
   &N/A
   \\ \hline\hline
\multicolumn{1}{|c|}{Pick-and-Place}                                                 & \begin{tabular}[c]{@{}c@{}}1.00\\ (0.00)\end{tabular}           & \begin{tabular}[c]{@{}c@{}}1.00\\ (0.00)\end{tabular}           & \begin{tabular}[c]{@{}c@{}}1.00\\ (0.00)\end{tabular}           & N/A           & \begin{tabular}[c]{@{}c@{}}1.00\\ (0.00)\end{tabular}           & N/A           \\ \hline
\end{tabular}
}
\caption{Number of executed actions in the planned path for various algorithms in simulation are presented here. Both \textsl{PokeRRT*} and  \textsl{Low-Impulse PokeRRT*} lead to fewer actions than their RRT-based counterparts, thus exploiting the large displacement property inherent to poking manipulation.\vspace{-12pt}}
\label{tab:eval-sim-actions}
\end{table}

\subsection{Simulation Experiments}

\cref{tab:eval-sim-time} shows the task times and success rates for \textsl{Pick-and-Place} and five non-prehensile manipulation planners---\textsl{PokeRRT*},  \textsl{Low-Impulse PokeRRT*},  \textsl{PokeRRT},  \textsl{Low-Impulse PokeRRT}, and \textsl{Two-Level Push Planner}. Results are averaged over 250 trials. \textsl{PokeRRT} and \textsl{PokeRRT*} successful in all scenarios while \textsl{Two-Level Push Planner} fails in S5 and S6 and grasping fails in S4 and S6. \textsl{Two-Level Push Planner} has a low overall success rate ($44\%$), as expected based on the reasons presented in \cref{sec:evaluation-scenarios}.

\textsl{PokeRRT} and \textsl{PokeRRT*} task times are lower than \textsl{Two-Level Push Planner} task times across all scenarios. Standard deviations are high due to the sampling-based nature of our planners. The task time for S4 is greater than for S1 since the object used in S4 is not only bigger in size but also larger in mass than the object used in S1, therefore poke displacements are lower given the same contact force. \textsl{Pick-and-Place} has the lowest task time because it does not involve planning in the object configuration space---the robot moves to object pose, grasps, and moves to goal pose, so only a single action is executed. Low-impulse poke planners have large search trees due to shorter displacements of the object. This results in longer planning times and also lower success rates. Task times are higher for obstacle scenarios because the robot end-effector is more likely to run into obstacles and collision checking is a computationally expensive procedure.

\textsl{Pick-and-Place} succeeds in simulation for S1, S2, S3, and S5 because there is no uncertainty in object pose so manipulation consists of moving to a predefined grasp configuration and moving to the goal pose. We intentionally set up the start and goal configurations for \textsl{Pick-and-Place} to create an upper bound for comparison with our planners. For the reasons presented in \cref{sec:evaluation-scenarios}, it fails in S4 and S6. Low-impulse poke planners have higher success rates than the push planner because the shorter robot control trajectories corresponding to poking result in fewer obstacle collisions. Low-impulse poke planners fail in S5 and S6 because the pokes are not strong enough to pass the workspace divider or cross into the second robot's reachable workspace.

% \begin{figure}
%   \centering
%     \includegraphics[width=\linewidth]{figures/sampleplans}
%   \caption{Plans generated by \textsl{PokeRRT} (a) involve more pokes than those generated by \textsl{PokeRRT*} (b), which indicates that \textsl{PokeRRT*} is better able to exploit poking's key characteristic of covering large distances.\vspace{-28pt}}\label{fig:plans}
% \end{figure}
% \vspace{-4.5pt}
\cref{tab:eval-sim-actions} presents the number of executed actions in the final planned path for multiple planning algorithms across several scenarios. It indicates that the number of executed actions is lower for \textsl{PokeRRT} and \textsl{PokeRRT*} compared to \textsl{Two-Level Push Planner}, showing that poke planners can displace the object further with fewer actions in less time. Notably, the number of actions is lowest for \textsl{PokeRRT*}, thereby supporting our claim that minimizing the number of poking actions exploits the large displacement property of poking manipulation.
% (\cref{fig:plans}).
However, the task time is similar to that of \textsl{PokeRRT} since \textsl{PokeRRT*} introduces the choose parent and rewiring steps to \textsl{PokeRRT}, which increases the number of actions sampled while planning. In general, \textsl{PokeRRT} is preferable in scenarios with obstacles---even though the number of pokes for \textsl{PokeRRT*} is lower (i.e., faster overall execution), a greater percentage of time is spent on resampling actions in \textsl{PokeRRT*}, most of which lead to object--obstacle collisions. Task times for low-impulse poke planners are comparable to \textsl{Two-Level Push Planner}, supporting our claim that pushing is a limiting case of poking where applied impulses are low in magnitude.

\begin{figure}
\vspace{4pt}
  \centering
    \includegraphics[width=\linewidth]{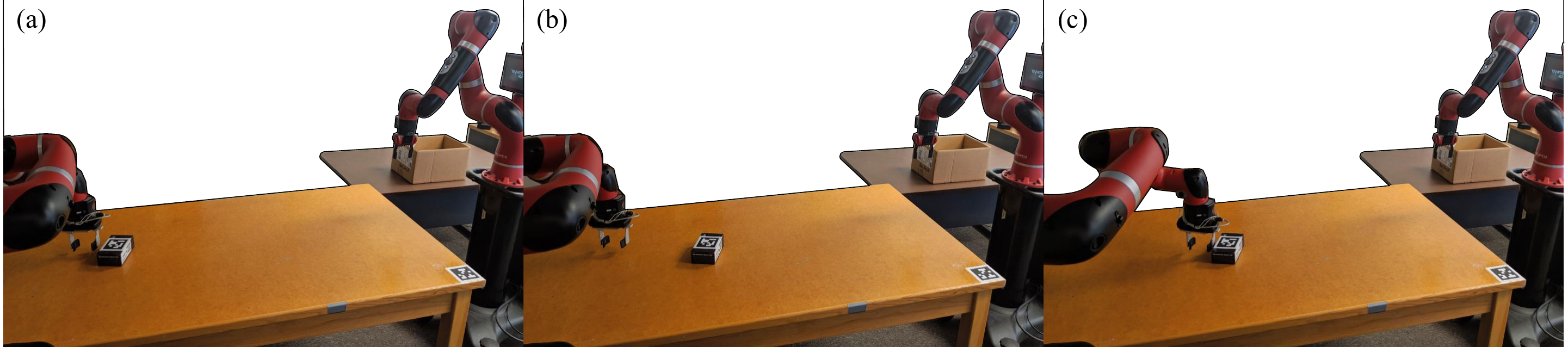}\\
    \includegraphics[width=\linewidth]{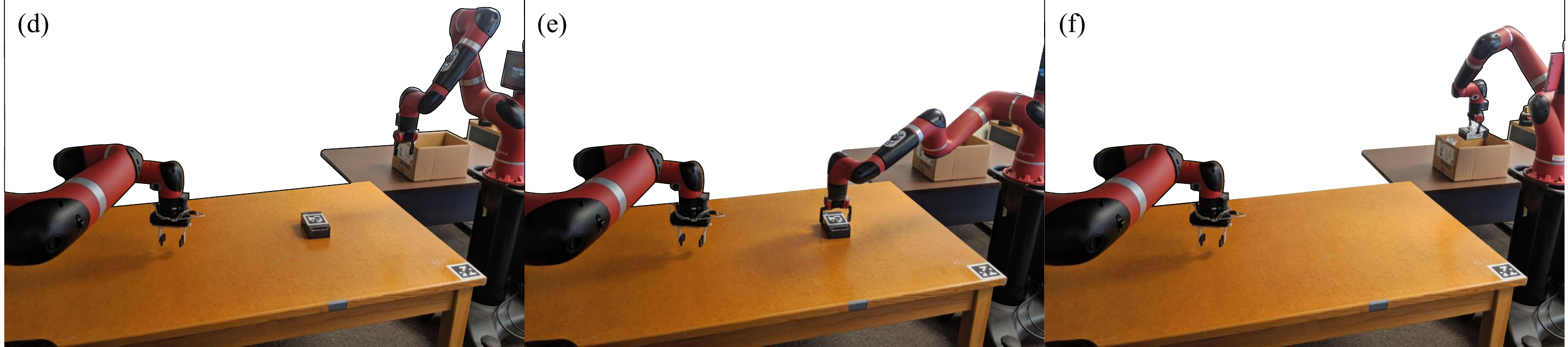}
  \vspace{-12pt}
  \caption{Two robots with non-overlapping reachable regions are shown (S6). Robot A (left) applies 2 pokes to manipulate the object to Robot B's (right) workspace (a-d). Robot B then grasps the object and places it in a bin that is not reachable by Robot A (e-f).\vspace{-15pt}}\label{fig:vignette}
\end{figure}

\subsection{Real-World Experiments}

\begin{figure}
\vspace{4pt}
  \centering
    \hspace{-7pt}\includegraphics[width=\linewidth]{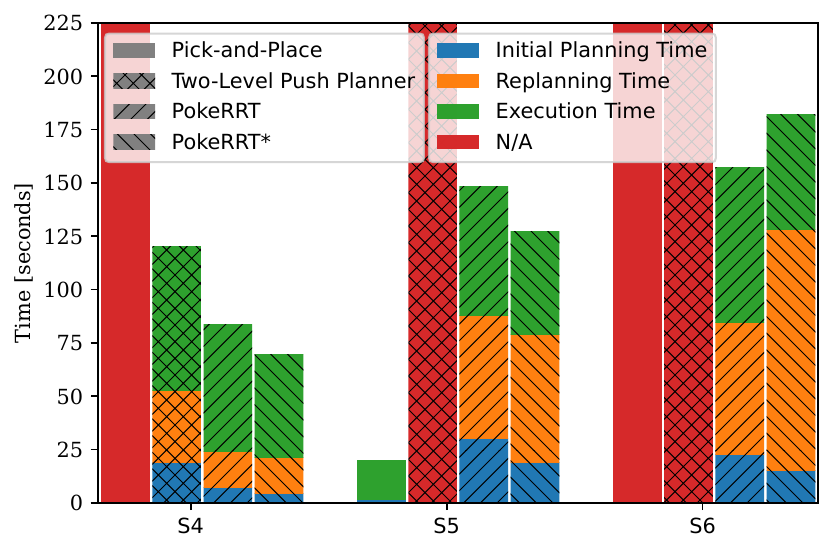}
%   \vspace{-12pt}
  \caption{A breakdown of task time as the sum of initial planning, replanning, and execution times is presented for various planners in the real-world. A single bar indicates total task time. Long red bars indicate cases where tasks cannot be solved. Overall, poke planning demonstrates lower execution and replanning times than push planning. \vspace{-10pt}}\label{fig:plot}
\end{figure}

We evaluate \textsl{PokeRRT*}, \textsl{PokeRRT}, \textsl{Two-Level Push Planner}, and \textsl{Pick-and-Place} for a subset of the scenarios in the real-world (\cref{tab:eval-real}). The task times and the number of actions executed in the real world are slightly higher than in simulation. This is expected because the plan is generated in simulation and simulation does not fully capture the complexities of the real-world environment. Therefore, replanning is required for real-world plan execution which leads to an increase in task times. As shown in \cref{fig:plot}, execution times for poke planners are lower than for the push planner indicating that poke planning leads to faster object manipulation than push planning. Additionally, the replanning time for push planning is higher than for poke planning since simulating push actions to generate the planning graph takes more computational cycles than simulating instantaneous contact for poke actions. However, the number of replans for push planning is lower than for poke planning thereby pointing to the inherently uncontrollable nature of poking, which may not be desirable in certain situations (\cref{tab:eval-real}).
Collectively, real-world results align with the results from simulation.
% Similar to simulation results, \textsl{PokeRRT*} has the fewest number of executed actions. In S4, poking is 1.7 times faster than pushing. In S5, pushing fails because of collision with obstacles and grasping fails because the object is too large for the gripper. Both pushing and grasping fail in S6 because the goal region is unreachable.

\cref{fig:vignette} depicts a failure recovery case for S6 where failure to grasp or push to the second robot's reachable workspace does not result in task failure---the first robot pokes the object to the second robot's reachable workspace, allowing the second robot to successfully manipulate the object.
Additionally, while grasping is a faster form of manipulation with fewer actions than poking or pushing, it has a lower overall success rate ($27\%$) than pushing ($33\%$) due to perception inaccuracies. This discrepancy between simulation and real-world results points to a fundamental difference between grasping and poking or pushing---sensing uncertainty leads to total failure in grasping, whereas for poking it leads to just partial failure as our presented planners operate in a closed-loop manner.

\begin{table*}[]
\vspace{4pt}
\centering\small
\resizebox{0.95\textwidth}{!}{
\centering
\begin{tabular}{|c|c|c|c|c|c|}
  \hline
   \textbf{Planner} & & PokeRRT* & PokeRRT & Two-Level Push Planner & Pick-and-Place \\
   \hline
  \multirow{3}{*}{\textbf{Task Time [seconds]}} & \textbf{S4} & \textbf{69.58 (26.15)} & 83.55 (29.59) & 120.21 (47.09) & N/A \\ \cline{2-6}
  & \textbf{S5} & \textbf{127.45 (57.80)} & 148.55 (86.23) & N/A & 20.31 (3.21) \\ \cline{2-6}
  & \textbf{S6} & 182.03 (122.77) & \textbf{157.03 (94.41)} & N/A & N/A \\ \hline
  \multirow{3}{*}{\textbf{Number of Actions}} & \textbf{S4} & \textbf{5.27 (2.00)} & 6.20 (1.60) & 6.00 (1.26) & N/A \\ \cline{2-6}
  & \textbf{S5} & 5.12 (1.05) & 6.67 (1.56) & N/A & \textbf{1.00 (0.00)} \\ \cline{2-6}
  & \textbf{S6} & \textbf{6.11 (1.91)} & 6.88 (2.15) & N/A & N/A \\ \hline
%   \multirow{3}{*}{\textbf{Initial Planning Time}} & \textbf{S4} & 4.64 (1.70) & 7.19 (7.26) & 19.10 (8.66) & N/A \\ \cline{2-6} 
%   & \textbf{S5} & 18.95 (12.41) & 29.97 (31.22) & N/A & N/A \\ \cline{2-6}
%   & \textbf{S6} & 15.38 (14.35) & 22.91 (34.66) & N/A & N/A \\ \hline
  \multirow{3}{*}{\textbf{Number of Replans}} & \textbf{S4} & 3.45 (1.67) & 4.2 (1.6) & \textbf{2.80 (1.54)} & N/A \\ \cline{2-6} 
  & \textbf{S5} & \textbf{3.38 (1.32)} & 4.11 (1.37) & N/A & N/A \\ \cline{2-6}
  & \textbf{S6} & \textbf{4.22 (1.81)} & 4.38 (1.49) & N/A & N/A \\ \hline
%   \multirow{3}{*}{\textbf{Replanning Times}} & \textbf{S4} & 4.75 (1.72) & 3.4 (2.01) & 10.04 (4.56) & N/A \\ \cline{2-6} 
%   & \textbf{S5} & 16.42 (9.89) & 11.58 (10.79) & N/A & N/A \\ \cline{2-6}
%   & \textbf{S6} & 22.56 (19.65) & 11.28 (10.32) & N/A & N/A \\ \hline
%   \multirow{3}{*}{\textbf{Execution Time}} & \textbf{S4} & 9.21 (0.5) & 9.53 (0.86) & 11.16 (0.46) & N/A \\ \cline{2-6} 
%   & \textbf{S5} & 9.47 (0.48) & 9.11 (0.28) & N/A & N/A \\ \cline{2-6}
%   & \textbf{S6} & 8.84 (0.23) & 10.20 (1.70) & N/A & N/A \\ \hline
  \textbf{Success Rate} & \textbf{S4 - S6} & \textbf{0.9 (0.3)} & \textbf{0.9 (0.3)} & 0.33 (0.47) & 0.27 (0.44)\\ \hline
\end{tabular}
}
\caption{Real-world task times, number of actions in executed path, and success rates are presented in this table. Results are averaged across 10 trials. Pushing has the highest success rate and fastest task times. PokeRRT* generates plans with the fewest number of actions.\vspace{-12pt}}
\label{tab:eval-real}
\end{table*}

\section{Conclusion and Discussion}\label{sec:discussion}

In this work, we demonstrate poking manipulation as a fundamental motion primitive that complements grasping and encompasses pushing in terms of capability. Our work is the first to show qualitative and quantitative results for multiple test conditions to demonstrate the flexibility and robustness of poking as a skill through \textsl{PokeRRT}. We present the task times, number of executed actions, and success rates of our proposed motion planners and four baseline algorithms across six different scenarios. The results demonstrate the strengths of the poking motion primitive: poking is not as limited by robot reachability, robot end-effector design, object properties, and inaccuracies due to perception as grasping or pushing. Success rates are higher for poking-based planners than for the push planner indicating that poking expands the size of reachable workspace by its ability to execute longer object displacements using shorter robot end-effector trajectories. Task times for computed plans are significantly lower for poking than for pushing, indicating that poking allows fast object manipulation because it does not face the same constant-contact restriction as pushing. \textsl{PokeRRT} and \textsl{PokeRRT*} also demonstrated the qualitative behaviors visualized in \cref{fig:scenarios}.

Errors in dynamic modeling caused by the gap between simulation and the real-world are covered by the replanning strategy in our work. However, future work in identifying sources of uncertainty will allow for the quantification of object--obstacle collision margins and incorporating them directly into the planning process will lower the chances of collision in the real-world. In order to make plan execution in the real-world more robust, our future work will also formalize analytical and learned models for poking and quantify object constraints which will lead to additional insights about the capabilities and limitations of poking. For instance, poking may not be an effective mode of manipulation for high friction interactions and should also be used sparingly since it may cause significant wear-and-tear on a real robot through repeated execution of high-speed trajectories. Consequently, by discovering the feasibility conditions of multiple skills (e.g. poking, pushing, and grasping) through interaction and planning around the key strengths of each motion primitive, robots will be able to more efficiently manipulate objects. 
\vspace{-2pt}

\bibliographystyle{IEEEtran}
\bibliography{icra2022}

% Generated by IEEEtran.bst, version: 1.14 (2015/08/26)
\begin{thebibliography}{10}
\providecommand{\url}[1]{#1}
\csname url@samestyle\endcsname
\providecommand{\newblock}{\relax}
\providecommand{\bibinfo}[2]{#2}
\providecommand{\BIBentrySTDinterwordspacing}{\spaceskip=0pt\relax}
\providecommand{\BIBentryALTinterwordstretchfactor}{4}
\providecommand{\BIBentryALTinterwordspacing}{\spaceskip=\fontdimen2\font plus
\BIBentryALTinterwordstretchfactor\fontdimen3\font minus
  \fontdimen4\font\relax}
\providecommand{\BIBforeignlanguage}[2]{{%
\expandafter\ifx\csname l@#1\endcsname\relax
\typeout{** WARNING: IEEEtran.bst: No hyphenation pattern has been}%
\typeout{** loaded for the language `#1'. Using the pattern for}%
\typeout{** the default language instead.}%
\else
\language=\csname l@#1\endcsname
\fi
#2}}
\providecommand{\BIBdecl}{\relax}
\BIBdecl

\bibitem{bullock2011classifying}
I.~M. Bullock and A.~M. Dollar, ``Classifying human manipulation behavior,'' in
  \emph{2011 IEEE International Conference on Rehabilitation Robotics}.\hskip
  1em plus 0.5em minus 0.4em\relax IEEE, 2011, pp. 1--6.

\bibitem{billard2019trends}
A.~Billard and D.~Kragic, ``Trends and challenges in robot manipulation,''
  \emph{Science}, vol. 364, no. 6446, 2019.

\bibitem{lynch1999dynamic}
K.~M. Lynch and M.~T. Mason, ``Dynamic nonprehensile manipulation:
  Controllability, planning, and experiments,'' \emph{The International Journal
  of Robotics Research}, vol.~18, no.~1, pp. 64--92, 1999.

\bibitem{Huang-1997-14452}
W.~Huang, ``Impulsive manipulation,'' Ph.D. dissertation, Carnegie Mellon
  University, Pittsburgh, PA, August 1997.

\bibitem{pasricha2021pokerrt}
A.~Pasricha, Y.~S. Tung, B.~Hayes, and A.~Roncone, ``{PokeRRT: A Kinodynamic
  Planning Approach for Poking Manipulation},'' in \emph{IROS 2021 Workshop on
  Impact-Aware Robotics}.\hskip 1em plus 0.5em minus 0.4em\relax IEEE, 2021.

\bibitem{ruggiero2018nonprehensile}
F.~Ruggiero, V.~Lippiello, and B.~Siciliano, ``Nonprehensile dynamic
  manipulation: A survey,'' \emph{IEEE Robotics and Automation Letters},
  vol.~3, no.~3, pp. 1711--1718, 2018.

\bibitem{toussaint2018differentiable}
M.~A. Toussaint, K.~R. Allen, K.~A. Smith, and J.~B. Tenenbaum,
  ``Differentiable physics and stable modes for tool-use and manipulation
  planning,'' 2018.

\bibitem{zeng2020tossingbot}
A.~Zeng, S.~Song, J.~Lee, A.~Rodriguez, and T.~Funkhouser, ``Tossingbot:
  Learning to throw arbitrary objects with residual physics,'' \emph{IEEE
  Transactions on Robotics}, vol.~36, no.~4, pp. 1307--1319, 2020.

\bibitem{woodruff2017planning}
J.~Z. Woodruff and K.~M. Lynch, ``Planning and control for dynamic,
  nonprehensile, and hybrid manipulation tasks,'' in \emph{2017 IEEE
  International Conference on Robotics and Automation (ICRA)}.\hskip 1em plus
  0.5em minus 0.4em\relax IEEE, 2017, pp. 4066--4073.

\bibitem{RSS2020Changkyu}
C.~Song and A.~Boularias, ``Learning to slide unknown objects with
  differentiable physics simulations,'' in \emph{Proceedings of Robotics:
  Science and Systems (RSS), Corvallis, Oregon}, 2020.

\bibitem{mason1986mechanics}
M.~T. Mason, ``Mechanics and planning of manipulator pushing operations,''
  \emph{The International Journal of Robotics Research}, vol.~5, no.~3, pp.
  53--71, 1986.

\bibitem{bauza2019omnipush}
M.~Bauza, F.~Alet, Y.-C. Lin, T.~Lozano-P{\'e}rez, L.~P. Kaelbling, P.~Isola,
  and A.~Rodriguez, ``Omnipush: accurate, diverse, real-world dataset of
  pushing dynamics with {RGB-D} video,'' in \emph{2019 IEEE/RSJ International
  Conference on Intelligent Robots and Systems (IROS)}.\hskip 1em plus 0.5em
  minus 0.4em\relax IEEE, 2019, pp. 4265--4272.

\bibitem{zito2012two}
C.~Zito, R.~Stolkin, M.~Kopicki, and J.~L. Wyatt, ``Two-level {RRT} planning
  for robotic push manipulation,'' in \emph{2012 IEEE/RSJ International
  Conference on Intelligent Robots and Systems}.\hskip 1em plus 0.5em minus
  0.4em\relax IEEE, 2012, pp. 678--685.

\bibitem{zhou2019pushing}
J.~Zhou, Y.~Hou, and M.~T. Mason, ``Pushing revisited: Differential flatness,
  trajectory planning, and stabilization,'' \emph{The International Journal of
  Robotics Research}, vol.~38, no. 12-13, pp. 1477--1489, 2019.

\bibitem{kloss2020accurate}
A.~Kloss, M.~Bauza, J.~Wu, J.~B. Tenenbaum, A.~Rodriguez, and J.~Bohg,
  ``Accurate vision-based manipulation through contact reasoning,'' in
  \emph{2020 IEEE International Conference on Robotics and Automation
  (ICRA)}.\hskip 1em plus 0.5em minus 0.4em\relax IEEE, 2020, pp. 6738--6744.

\bibitem{li2018push}
J.~K. Li, W.~S. Lee, and D.~Hsu, ``{Push-Net}: Deep planar pushing for objects
  with unknown physical properties.'' in \emph{Robotics: Science and Systems},
  2018.

\bibitem{king2015nonprehensile}
J.~E. King, J.~A. Haustein, S.~S. Srinivasa, and T.~Asfour, ``Nonprehensile
  whole arm rearrangement planning on physics manifolds,'' in \emph{2015 IEEE
  International Conference on Robotics and Automation (ICRA)}.\hskip 1em plus
  0.5em minus 0.4em\relax IEEE, 2015, pp. 2508--2515.

\bibitem{haustein2015kinodynamic}
J.~A. Haustein, J.~King, S.~S. Srinivasa, and T.~Asfour, ``Kinodynamic
  randomized rearrangement planning via dynamic transitions between statically
  stable states,'' in \emph{2015 IEEE International Conference on Robotics and
  Automation (ICRA)}.\hskip 1em plus 0.5em minus 0.4em\relax IEEE, 2015, pp.
  3075--3082.

\bibitem{8207585}
Muhayyuddin, M.~Moll, L.~Kavraki, and J.~Rosell, ``Randomized physics-based
  motion planning for grasping in cluttered and uncertain environments,''
  \emph{IEEE Robotics and Automation Letters}, vol.~3, no.~2, pp. 712--719,
  2018.

\bibitem{zhu2005frictional}
C.~Zhu, Y.~Aiyama, T.~Arai, and A.~Kawamura, ``Frictional sliding motion in
  releasing manipulation,'' \emph{Advanced Robotics}, vol.~19, no.~2, pp.
  141--168, 2005.

\bibitem{coumans2019}
E.~Coumans and Y.~Bai, ``{PyBullet}, a {Python} module for physics simulation
  for games, robotics and machine learning,'' \url{http://pybullet.org},
  2016--2019.

\bibitem{collins2019quantifying}
J.~Collins, D.~Howard, and J.~Leitner, ``Quantifying the reality gap in robotic
  manipulation tasks,'' in \emph{2019 International Conference on Robotics and
  Automation (ICRA)}.\hskip 1em plus 0.5em minus 0.4em\relax IEEE, 2019, pp.
  6706--6712.

\bibitem{7363472}
P.~{Beeson} and B.~{Ames}, ``{TRAC-IK}: An open-source library for improved
  solving of generic inverse kinematics,'' in \emph{2015 IEEE-RAS 15th
  International Conference on Humanoid Robots (Humanoids)}, 2015, pp. 928--935.

\bibitem{lavalle1998rapidly}
S.~M. LaValle, ``Rapidly-exploring random trees: A new tool for path
  planning,'' 1998.

\bibitem{karaman2011sampling}
S.~Karaman and E.~Frazzoli, ``Sampling-based algorithms for optimal motion
  planning,'' \emph{The International Journal of Robotics Research}, vol.~30,
  no.~7, pp. 846--894, 2011.

\bibitem{karaman2011anytime}
S.~Karaman, M.~R. Walter, A.~Perez, E.~Frazzoli, and S.~Teller, ``Anytime
  motion planning using the rrt,'' in \emph{2011 IEEE International Conference
  on Robotics and Automation}.\hskip 1em plus 0.5em minus 0.4em\relax IEEE,
  2011, pp. 1478--1483.

\end{thebibliography}

\end{document}